\documentclass[letterpaper, 10 pt, journal, twoside]{ieeetran}  

\IEEEoverridecommandlockouts                              

\usepackage{graphicx} 
\usepackage{subfigure}  
\usepackage{overpic}  
\usepackage{float}
\usepackage{stfloats}
\usepackage{tabu} 
\usepackage{multirow}
\usepackage{makecell}    
\usepackage{array}
\usepackage{amsfonts}
\usepackage{amsmath, amssymb, mathtools}
\usepackage{algorithm}
\usepackage{algpseudocode}
\usepackage[noadjust]{cite}
\usepackage[normalem]{ulem}
\usepackage[colorlinks,linkcolor=red]{hyperref}
\usepackage{balance}

\usepackage{enumitem}

\title{Part-Guided 3D RL for Sim2Real Articulated Object Manipulation}

\author{Pengwei Xie$^{*1,5}$, Rui Chen$^{*3}$, Siang Chen$^{*1,2,5}$, Yuzhe Qin$^{4}$, Fanbo Xiang$^{4}$, Tianyu Sun$^{1,5}$, \\ Jing Xu$^{3}$, Guijin Wang$^{\dagger 1,2,5}$, Hao Su$^{4}$ 
\thanks{$^{1}$Department of Electronic Engineering, Tsinghua University, Beijing 100084, China.}
\thanks{$^{2}$Shanghai AI Laboratory, Shanghai 200232, China.}
\thanks{$^{3}$Department of Mechanical Engineering, Tsinghua University, Beijing 100084, China.}
\thanks{$^{4}$Department of Computer Science, University of California San Diego.}
\thanks{$^{5}$Beijing National Research Center for Information Science and Technology (BNRist), China.}
\thanks{$^*$Equal Contribution.}
\thanks{$\dagger$Correspondence: {\tt wangguijin@tsinghua.edu.cn.}}
}

\begin{document}

\bstctlcite{setting}


\maketitle

\begin{abstract}
Manipulating unseen articulated objects through visual feedback is a critical but challenging task for real robots. Existing learning-based solutions mainly focus on visual affordance learning or other pre-trained visual models to guide manipulation policies, which face challenges for novel instances in real-world scenarios. In this paper, we propose a novel part-guided 3D RL framework, which can learn to manipulate articulated objects without demonstrations. We combine the strengths of 2D segmentation and 3D RL to improve the efficiency of RL policy training. To improve the stability of the policy on real robots, we design a Frame-consistent Uncertainty-aware Sampling (FUS) strategy to get a condensed and hierarchical 3D representation. In addition, a single versatile RL policy can be trained on multiple articulated object manipulation tasks simultaneously in simulation and shows great generalizability to novel categories and instances. Experimental results demonstrate the effectiveness of our framework in both simulation and real-world settings. Our code is available at \href{https://github.com/THU-VCLab/Part-Guided-3D-RL-for-Sim2Real-Articulated-Object-Manipulation}{https://github.com/THU-VCLab/Part-Guided-3D-RL-for-Sim2Real-Articulated-Object-Manipulation}
\end{abstract}


\section{INTRODUCTION}

\IEEEPARstart{A}{rticulated} object manipulation is a fundamental problem in robotics. 
Unlike object grasping, which only requires stable contact between the gripper and the object~\cite{qin2020s4g, fang2020graspnet}, articulated objects manipulation involves controlling the relative motion of their parts and is more challenging due to the complex kinematic structures and dynamic properties.

Recently, learning-based methods have emerged as a promising approach to learning manipulation policies from visual inputs \cite{mo2021where2act, wu2022vatmart, geng2022end, xu2022universal, EisnerZhang2022FLOW, ma2023sim2real2, jiang2022ditto}. Some approaches, such as \cite{jiang2022ditto, ma2023sim2real2}, reconstruct the object and estimate its kinematic and dynamic properties, but require interactive perception and are not practical for novel objects. Other methods \cite{mo2021where2act, wu2022vatmart, geng2022end, xu2022universal, EisnerZhang2022FLOW} learn visual affordance information to guide the manipulation policies. However, these methods still face several limitations, such as the need for expert knowledge to control robot motion, the low sample efficiency of reinforcement learning (RL) algorithms, and the significant shape variations of articulated objects within and across different categories. Furthermore, the generalizability and robustness of these methods on real robots have yet to be sufficiently evaluated.

\begin{figure}[tbp] 
\centering
\includegraphics[width=0.45\textwidth]{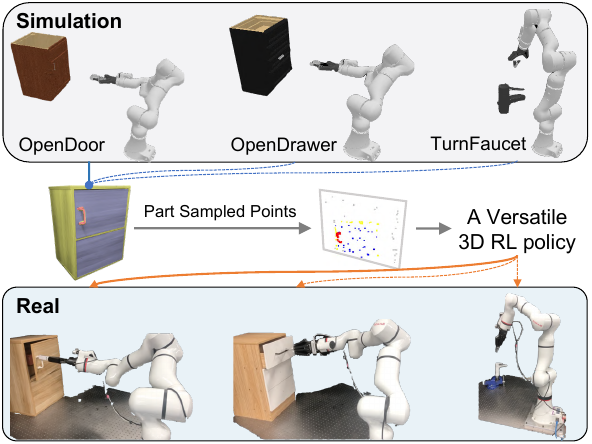}
\caption{\textcolor{black}{Our 3D RL framework} can be trained for various articulated object manipulation tasks in simulation simultaneously and efficiently. After training, without any demonstrations, the versatile RL policy can be deployed to a real robot and perform different tasks.}
\label{Fig.Sim2RealTasks}
\end{figure}

In this work, we propose an efficient part-guided 3D RL framework for articulated object manipulation without any demonstrations, as shown in Fig. \ref{Fig.Sim2RealTasks}. First, we train a part segmentation module on synthetic data and implement it to segment the 2D image into different articulated parts. Then, we transform the segmentation result into 3D points with part information, as 3D representations are more appropriate for reasoning the relationship between the robot and the object. Finally, we extract geometric features from the points to guide the RL policy training. By leveraging the strength of both the 2D segmentation network and 3D RL, our framework not only improves the sample efficiency but also enables the training of a single versatile RL policy capable of handling manipulation tasks for various articulated objects with different kinematic structures.

In particular, we design a Frame-consistent Uncertainty-aware Sampling (FUS) strategy to obtain a condensed and hierarchical 3D object representation. Due to the Sim2Real gap and the variance of objects and illuminations in real-world scenarios, the 2D segmentation network trained on synthetic data cannot yield satisfactory results even with domain randomization. Therefore, we introduce a method to estimate the uncertainty of the segmentation predictions and analyze the consistency of points across consecutive frames to obtain a more reliable and accurate object representation. Compared with traditional sampling methods, our sampling method preserves the object structure with higher fidelity, resulting in a more stable manipulation policy on real robots.

The key contributions of our study are as follows.
\begin{enumerate}
    \item[(1)] We propose a generic framework for articulated object manipulation, which exploits the advantages of 2D segmentation and 3D RL. It enables learning a versatile policy to handle different articulated objects.
    \item [(2)] We design a novel weighted point sampling strategy considering both point uncertainty and consistency, which ensures a condensed object representation and reduces the Sim2Real gap during policy deployment.
    \item [(3)] We conduct various quantitative evaluations in three articulated object manipulation tasks, both in simulation and real-world experiments, proving the effectiveness of our framework. 
\end{enumerate}

\section{RELATED WORK}
\subsection{Learning-based Articulated Objects Manipulation}
Learning-based methods have shown remarkable progress in vision-based articulated object manipulation. Affordance learning has been proposed for articulated objects, such as \cite{mo2021where2act, wu2022vatmart, wang2022adaafford, geng2022end, xu2022universal, EisnerZhang2022FLOW}. \textcolor{black}{Recent impressive works, such as UMPNet \cite{xu2022universal} and FlowBot3D \cite{EisnerZhang2022FLOW}, first learn visual affordance and then use human-designed motion commands for execution with suction grippers. \textcolor{black}{However, they may necessitate additional endeavors to address the grasping issue for other grippers.} In this paper, our RL approach works in an end-to-end fashion, demonstrating excellent performance in simulation and zero-shot transfer to the real world.}

\textcolor{black}{\cite{urakami2019doorgym, wangLearningSemanticKeypoint2020a} estimate one or several keypoints from images to guide the manipulation policy. In contrast, our method utilizes part-guided sampled points, which significantly enhances robustness and leads to better generalizability on novel object instances in both simulation and real world.}

\subsection{Visual Pre-training for Robot Learning}
Recent research has demonstrated the effectiveness of visual pre-training in improving robot learning \cite{yen2020learning, radosavovic2022realworld, seo2022reinforcement, zhan2022learning, ma2023vip, xiao2022maskedvpt}. Lin et al. \cite{yen2020learning} have shown that visual representations from semantic tasks highly correlate with affordance maps widely used in manipulation. Many recent works have leveraged self-supervised visual pre-training models on images to guide manipulation tasks \cite{radosavovic2022realworld, zhan2022learning, xiao2022maskedvpt}. However, most existing approaches \cite{seo2022reinforcement, ma2023vip} require learning from large-scale, diverse, offline human videos, which is computationally expensive and labor intensive. Furthermore, additional real demonstrations are often needed to transfer the learned policies to real-world robot tasks \cite{zhan2022learning, ma2023vip}. In contrast, our approach pre-trains the visual representations solely on synthetic data, which can be easily collected on a large scale. By incorporating domain randomization methods, our visual model achieves excellent performance on real captured images. Moreover, we lift the 2D representation to 3D sampled points, which further improves the training efficiency of RL.

\section{METHOD}
\subsection{Overview} \label{Method.Framework}

\begin{figure*}[tbp] 
\centering
\includegraphics[width=0.9\textwidth]{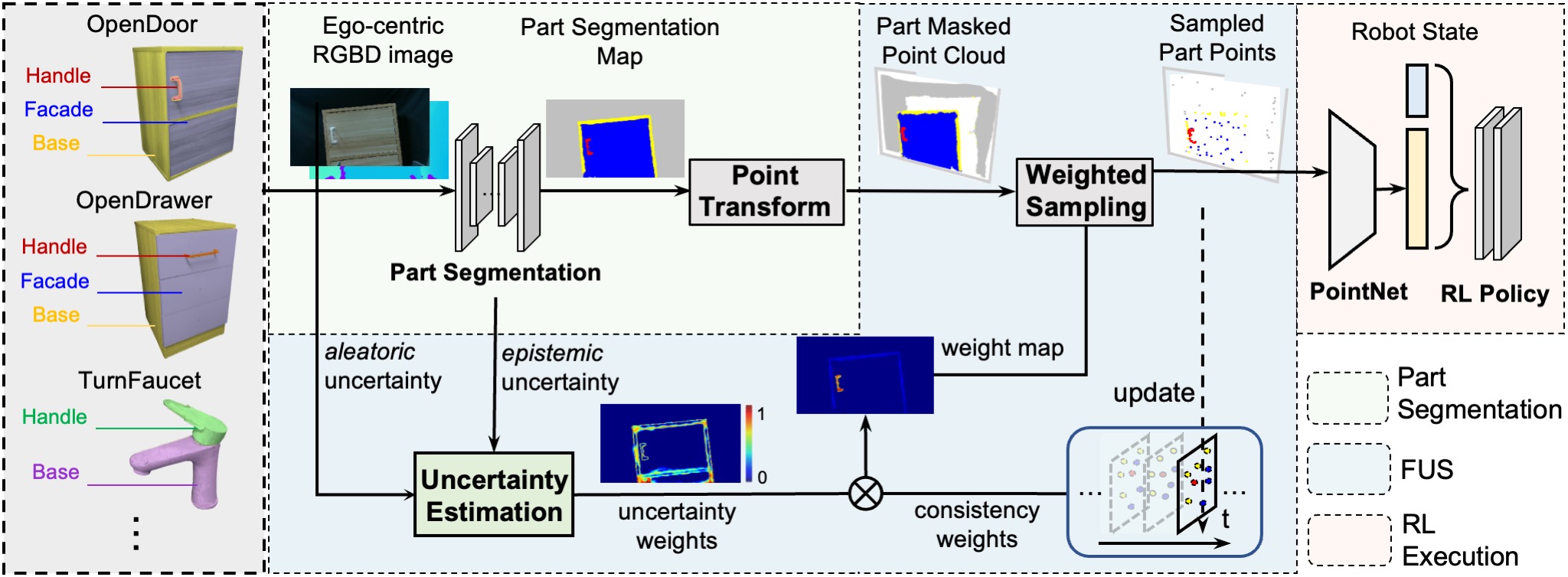}
\caption{\textcolor{black}{\textbf{Framework Overview}.} 1) We take hand-centric visual observation (i.e., RGB-D images) as input and predict the part segmentation map using a pre-trained segmentation network. 2) 3D part masked points are transformed from the depth image using the camera parameters. 3) Our proposed FUS strategy combines uncertainty and consistency weights to generate per-point weights. These weights are used to sample points for each part. 4) Geometric features extracted using PointNet are combined with the robot states, and fed to the RL algorithm to get the action. After the robot executes the action, a new observation is obtained, and the process iterates from the beginning. }
\label{Fig.Framework}
\end{figure*}

We model the articulated object manipulation problem as a Partially Observable Markov Decision Process (POMDP) \cite{kaelbling1998planning}, aiming to learn a policy $\pi:\mathcal{O}\rightarrow \mathcal{A}$. The observation $o \in \mathcal{O}$ includes visual features and robot states, and the action $a \in \mathcal{A}$ indicates the robot target joint positions and the gripper finger position. \textcolor{black}{In our work, the agent employs a hand-centric camera to mitigate occlusion issues from third-person perspectives, showing superior generalizability for manipulation tasks in both simulation and real world \cite{hsu2022visionbased}.}

As depicted in Fig. \ref{Fig.Framework}, our framework develops an efficient RL policy for various articulated object manipulation tasks. \textcolor{black}{Firstly, we take a hand-centric RGB-D image $\mathbf{I}\in \mathbb{R}^{H\times W\times 4}$ as input. Next, our segmentation model predicts the part segmentation map $\mathbf{S}=\{0,1\}^{C \times H \times W}$, where $C$ denotes the number of part classes. Our Frame-consistent Uncertainty-aware Sampling (FUS) strategy selects points based on the segmentation map. Finally, geometric features are extracted from these points and fed into the RL algorithm with robot states to predict actions and corresponding critic values.}

\subsection{Part Segmentation} \label{Method.SegmentModel}
\textcolor{black}{We aim to train an efficient part segmentation model from RGB-D images. Note that parts can also be segmented from point clouds. However, in our hand-centric setting, depth sensor noise may affect the quality of point clouds, especially when the camera is close to the object. Additionally, point cloud downsampling is necessary to enable efficient RL training. This decreases the points from small parts and degrades the segmentation performance potentially.}

\textcolor{black}{\textbf{Part Definition}: Our goal is to develop an object-irrelevant part representation that exhibits generalizability in manipulation tasks across various articulated objects. Similar to GAPartNet \cite{geng2023gapartnet}, our work defines a part as a rigid segment that shares similar affordances, facilitating generalizable and consistent interaction behavior. As shown in Fig. \ref{Fig.Framework}, cabinet doors and drawers consist of three parts: fixed handles, door/drawer facades, and fixed bases. Faucets are composed of handles and fixed bases. By adopting the generalizable part representation, our part-based RL method exhibits the potential to manipulate various types of articulated objects.}

\textbf{Synthetic Data Generation}: To generate a large amount of annotated data, we collect data from various categories of articulated objects in the simulation. First, We centrally position each object on a table and deploy a floating gripper fitted with a mounted camera to capture RGB-D images from diverse viewpoints. Corresponding part masks are automatically generated from the simulation without human intervention. And then, we introduce scene-level randomization to enrich the dataset, including variations in the distance between the gripper and object, camera orientation, turning angle of the articulated parts, and opening distance.

\textbf{Domain Randomization}: Domain randomization techniques have demonstrated great potential in bridging the gap between synthetic and real data by increasing the data's diversity \cite{tobin2017domain, urakami2019doorgym, wangLearningSemanticKeypoint2020a}. In our data generation process, we randomize several material parameters for the object mesh models, such as metallic, roughness, and specular parameters. Additionally,  random texture patterns are introduced on the surface of different objects. To further diversify the background patterns, we replace the original black background with random RGB-D images from SUNRGBD \cite{song2015sun} and SceneNetRGBD \cite{mccormac2017scenenet}. The introduction of large variations during data generation effectively expands the distribution of synthetic data. Several data augmentation techniques are also employed in the training phase to enhance the generalizability of our segmentation network. These techniques include color jittering for RGB images, adding salt-and-pepper noise to depth images, applying random cropping and rotation, and introducing Gaussian noise. We also simulate depth sensor noise during data generation to narrow the domain gap, as previous studies have shown its effectiveness \cite{zhang2022close}.

With the part masks, there are multiple approaches to guide the subsequent RL policy learning. \textcolor{black}{Many image-based RL approaches \cite{yarats2021image, yen2020learning, hsu2022visionbased} use an image encoder to extract image features.} However, 2D features may not be sufficient for 3D robotic manipulation tasks that require reasoning about the geometric structure and agent-object relationships in 3D space \cite{fang2020graspnet, liu2022frame}. Therefore, we opt for using 3D point cloud representations. Depth pixels are transformed to 3D points in the world frame using the calibrated camera's intrinsic and extrinsic parameters. As the original point clouds can be quite large, we apply downsampling methods to accelerate RL training. To ensure the reliability and consistency of the sampled points, we introduce a Frame-consistent Uncertainty-aware Sampling strategy.

\subsection{Frame-consistent Uncertainty-aware Sampling}
There are various sampling strategies for point clouds. Uniformly Downsampling is the most common approach \cite{mo2021where2act, wang2022adaafford, liu2022frame, qin2023dexpoint}. This method first filters out table points from the point cloud in the world frame, and then uniformly downsamples the remaining points to a fixed number. However, many points of key parts may be dropped, especially when the part is small. Farthest point sampling (FPS) is a commonly employed strategy in point cloud neural processing systems \cite{qi2019deep}. However, FPS does not consider subsequent processing steps on the sampled points and may therefore yield sub-optimal performance. To maintain the versatility of our segmentation model for manipulation tasks, we do not assume the significance of any specific part. Therefore, we aim to sample the same number of points from each part, namely $N_s$. For optimal results, points sampled from the same mask should accurately represent its shape information and display consistency throughout different stages of the manipulation policy. Therefore, our proposed weighted sampling strategy consists of uncertainty weights and consistency weights.

\textbf{Uncertainty Weights}: Although our segmentation model performs well in simulation, its performance may decrease in real-world scenarios due to domain gaps between simulation and reality, even though we have implemented various domain randomization techniques. Scattered and inconsistent sampled points can result from sub-optimal segmentation outcomes, impeding the learning of subsequent policies. Inspired by \cite{kendall2017uncertainties, wang2019aleatoric}, we employ the uncertainty estimation method to measure the potential uncertainty in the Sim2Real transfer process. Specifically, we utilize Test-Time Augmentation (TTA, such as flipping, color jittering on RGB images, and salt-and-pepper noise on depth images) to estimate \textit{aleatoric uncertainty} (inherent observation noise), and employ Monte Carlo Dropout \cite{kendall2017uncertainties} to estimate \textit{epistemic uncertainty} (model uncertainty).

First, given the input image $\mathbf{I}$ of the current time step, we compute the corresponding points $\mathbf{p}$ in the world frame by transforming the pixel coordinates $(\mathbf{u}, \mathbf{v})$ and depth values $\mathbf{d}$ using the calibrated camera intrinsic and extrinsic parameters. Next, we add TTA on $\mathbf{I}$ and perform $K$ stochastic forward inferences under random dropout. Therefore, we obtain a set of softmax probability map $\{\mathbf{P}_{k}$  \textcolor{black}{$\in \mathbb{R}^{C\times H\times W}$}$\}_{k=1}^{K}$. And then, we use the predictive entropy to approximate the uncertainty \cite{wang2019aleatoric}, summarized as
\begin{equation}
\textcolor{black}{\mathbf{P}_c=\frac{1}{K}\sum_k \mathbf{P}^c_{k}\ and\ 
\mathbf{U}=-\sum_c \mathbf{P}_c\log \mathbf{P}_c,}
\end{equation}
where $\mathbf{P}_{c}$ is the probability in the $c$-th part class at the current time step. And the uncertainty map $\mathbf{U}$\textcolor{black}{$\in \mathbb{R}^{H\times W}$} is then normalized to $[0,1]$. This uncertainty map allows us to assign different sampling weights to each point. Specifically, given \textcolor{black}{the part segmentation map $\mathbf{S}$} and the related normalized uncertainty score $\mathbf{U}$, the \textcolor{black}{part-based uncertainty and} uncertainty sampling weights can be calculated as
\begin{equation}
\textcolor{black}{\mathbf{U}_{c} = \big [\mathbf{U}_{(u,v)} \big | \mathop{\arg \max}\mathbf{S}_{(u,v)}=c \big ], c \in \{1,...,C\},}
\end{equation}
\begin{equation}
\label{eq.w_uncertainty}
\mathbf{w}_{c}^{ua} = \mathrm{softmax}(\mathbf{U}_{c}), \textcolor{black}{c \in \{1,...,C\}},
\end{equation}
where \textcolor{black}{$\mathbf{U}_{c}\in \mathbb{R}^{N_c\times 1}$ denotes the uncertainty score vector of pixels belonging to part $c$ and $N_c$ is the number of pixels. We apply the $\mathrm{softmax}$ function to obtain the normalized sampling weights for each part.}

\textbf{Consistency Weights}: The consistency of sampled points across frames is another key factor to consider. During the manipulation process, the points in the world frame often remain stable across consecutive frames. Even after the articulated part is moved, the points in the adjacent regions will not change significantly. Based on the above observation, we design the frame-consistent sampling weights to ensure the sampled points remain consistent in the process. First, for the initial frame, we randomly sample $N_s$ points from each part. Then, a queue $\mathcal{Q}$ of length $T_{fc}$ is kept to store the sampled points in the manipulation process. For the current time step, we compute the distance vector \textcolor{black}{$\mathbf{d}_{c} \in \mathbb{R}^{N_c\times 1}$ between the part points $\mathbf{p}_{c}\in \mathbb{R}^{N_c\times 3}$ and the part points $\mathbf{q}_{c}\in \mathbb{R}^{N_s\times 3}$} from $\mathcal{Q}$ for each part:
\begin{equation}
\label{eq.pts_distance}
\textcolor{black}{\mathbf{d}_{c} = \big [\mathop{\min}_{\mathbf{q}_j\in \mathbf{q}_c} \|\mathbf{p}_i -\mathbf{q}_j\| \big | \mathbf{p}_i\in \mathbf{p_{c}} \big ], c \in \{1,...,C\}.}
\end{equation}

\begin{figure}[tbp] 
\centering
\includegraphics[width=0.46\textwidth]{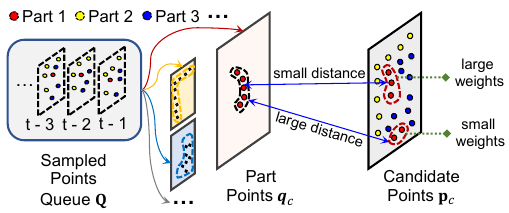}
\caption{We calculate consistency weights from several consecutive frames. Candidate points closer to former sampled ones from the same part are allocated larger sampling weights.}
\label{Fig.FrameConsist}
\end{figure}

As shown in Fig. \ref{Fig.FrameConsist}, noticing that the points from the same part are closer, and the noisy points are unstable and usually have much larger distances to the points of the specific part, we use the distance \textcolor{black}{$\mathbf{d}_{c}$ to calculate the frame-consistent sampling weights $\mathbf{w}_{c}^{fc}$ as}
\begin{equation}
\label{eq.w_consist}
\mathbf{w}_{c}^{fc} = 2^{-K^{fc} \cdot \textcolor{black}{\mathbf{d}_{c}}}, 
\end{equation}
where $K^{fc}$ denotes the decay coefficient.

\textcolor{black}{\textbf{Frame-consistent Uncertainty-aware Weights}:} Now we have uncertainty weights and consistency weights for each part, and the total sampling weights are calculated as
\begin{equation}
\label{eq.w_fcuw}
\mathbf{w}_{c}=\mathbf{w}_{c}^{fc}\circ \mathbf{w}_{c}^{ua}, \textcolor{black}{c \in \{1,...,C\}},
\end{equation}
where $\circ$ denotes element-wise multiplication. Then we sample $N_s$ points \textcolor{black}{$\mathbf{\hat{p}}_{c}$} from each part according to $\mathbf{w}_{c}$. \textcolor{black}{Then the combined part points $\mathbf{\hat{p}}$} are put into $\mathcal{Q}$, accounting for the calculation of $\mathbf{w}^{fc}_{c}$ of the next frame.

\textcolor{black}{As depicted in Fig. \ref{Fig.FrameConsist}, the consistency weights can be recovered when erroneous part points are removed from the queue. Furthermore, in our real experiments, we observe that the presence of individual frames with inaccurate segmentation does not significantly affect the subsequent steps.}

\subsection{RL Policy Learning}  \label{Method.RL}
\textcolor{black}{As Algorithm \ref{Alg.PartRL} shows, an RGB-D image $\mathbf{I}$ is first segmented into multiple parts. Then, we transform these parts into 3D points $\mathbf{p}$, incorporating their corresponding one-hot part-belonging encoding. Next, we conduct part-guided weighted sampling on the raw part points to acquire consistent part representation $\mathbf{\hat{p}}_c$. Similar to \cite{liu2022frame, wu2022vatmart, geng2022end}, a PointNet \cite{qi2017pointnet} is utilized to extract compact geometric features $\hat{F}$ from the combined part points $\mathbf{\hat{p}}$. Consequently, $\hat{F}$ is concatenated with the robot states $g$ and fed into the RL algorithm.}

\begin{algorithm}[htbp]
\caption{\textcolor{black}{Part-guided articulation manipulation policy}}
\label{Alg.PartRL}
\begin{algorithmic}
\Require{$\theta \gets$ parameters of the segmentation model $f_\theta$}
\While{not $EpisodeComplete()$}
\State $\mathbf{I} \gets$ the RGB-D image.
\State $\mathbf{S} \gets f_\theta(\mathbf{I})$, Predict the segmentation map.
\State $\mathbf{p} \gets PointTransform(\mathbf{I}, \mathbf{S})$.
\State Get $\mathbf{w}_{c}$ from Eq. \ref{eq.w_fcuw}.
\State $\mathbf{\hat{p}}:\bigcup_c\mathbf{\hat{p}}_c \gets WeightSampling(\mathbf{p}, \mathbf{w}_c)$
\State $\hat{F} \gets PointNet(\mathbf{\hat{p}})$, Get geometric features.
\State $a \gets Actor(\hat{F} \oplus g)$, Predict action.
\EndWhile
\end{algorithmic}
\end{algorithm}

\textbf{Reward Shaping}: \textcolor{black}{To effectively train the RL policy for common articulated object manipulation, we design a versatile reward function, which includes five standard terms. 1) Approaching term: encouraging the gripper to approach the movable part. 2) Direction term: promoting manipulation of the movable part in the appropriate direction. 3) Position term: driving the gripper to move the movable part to the target position. 4) Visibility term: encouraging the agent to maintain visual contact with the movable part. 5) Grasp term: encouraging the gripper to grasp the handles. All terms are utilized for \textit{OpenDoor} and \textit{OpenDrawer}. However, for the \textit{TurnFaucet} task, the Grasp term is removed since grasping is not necessary for manipulating faucets.}

\section{EXPERIMENTS}
To evaluate the performance of our proposed framework in both simulated and real-world scenarios, we aim to answer the following research questions: (1) How does our method compare to existing methods for articulated object manipulation tasks? (2) How effective is our framework for real-world robot applications, particularly in addressing the Sim2Real gap? (3) How does our part-based sampling strategy compare to other traditional methods regarding accuracy and stability? (4) To what extent is the versatile RL policy trained on three tasks simultaneously effective in both simulation and real-world experiments?

\begin{figure}[tbp]
\centering
    \subfigure[Real Setup]{
        \includegraphics[width=0.35\linewidth]{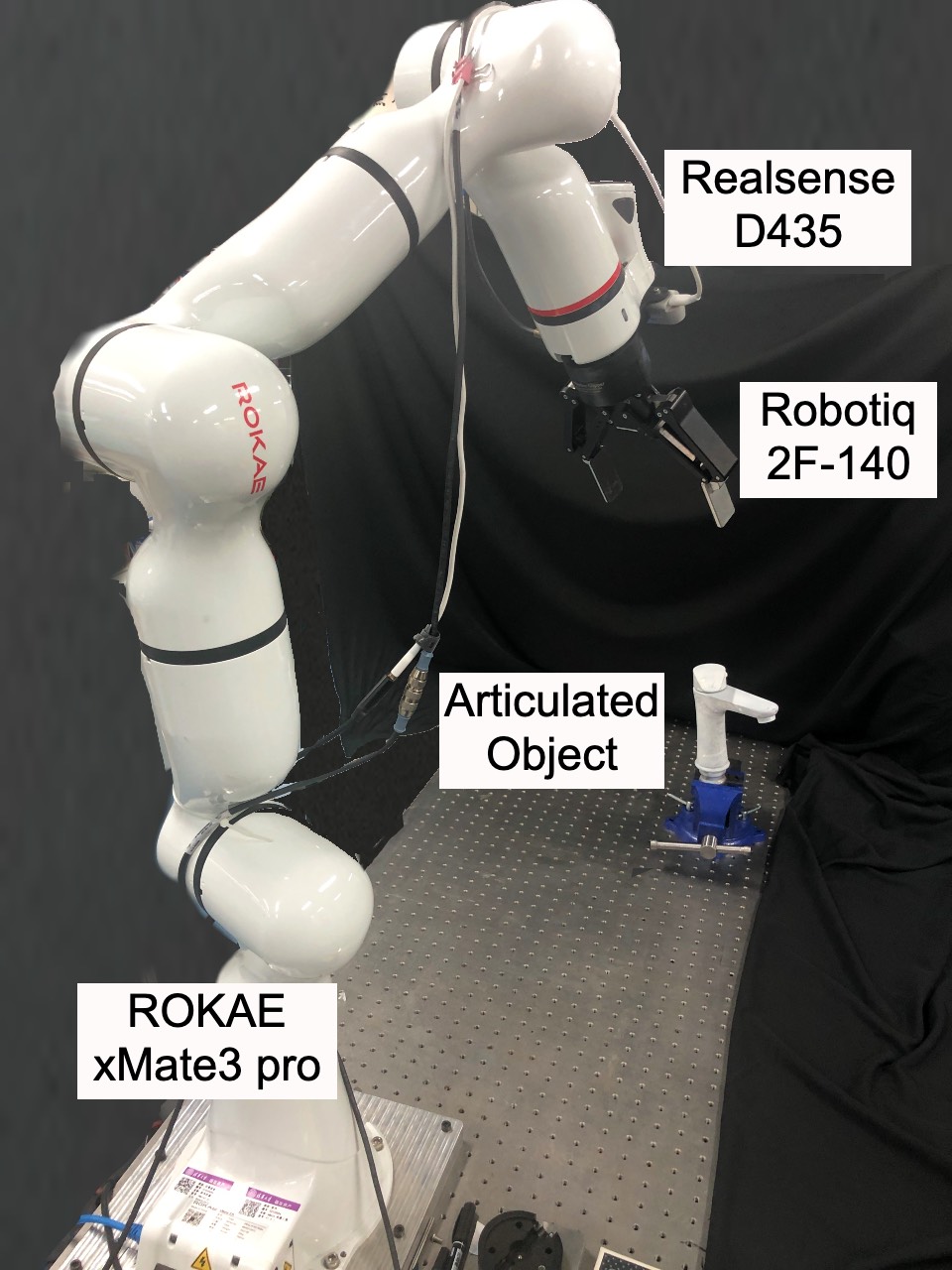}
    }
    \subfigure[Real articulated objects]{
        \includegraphics[width=0.35\linewidth]{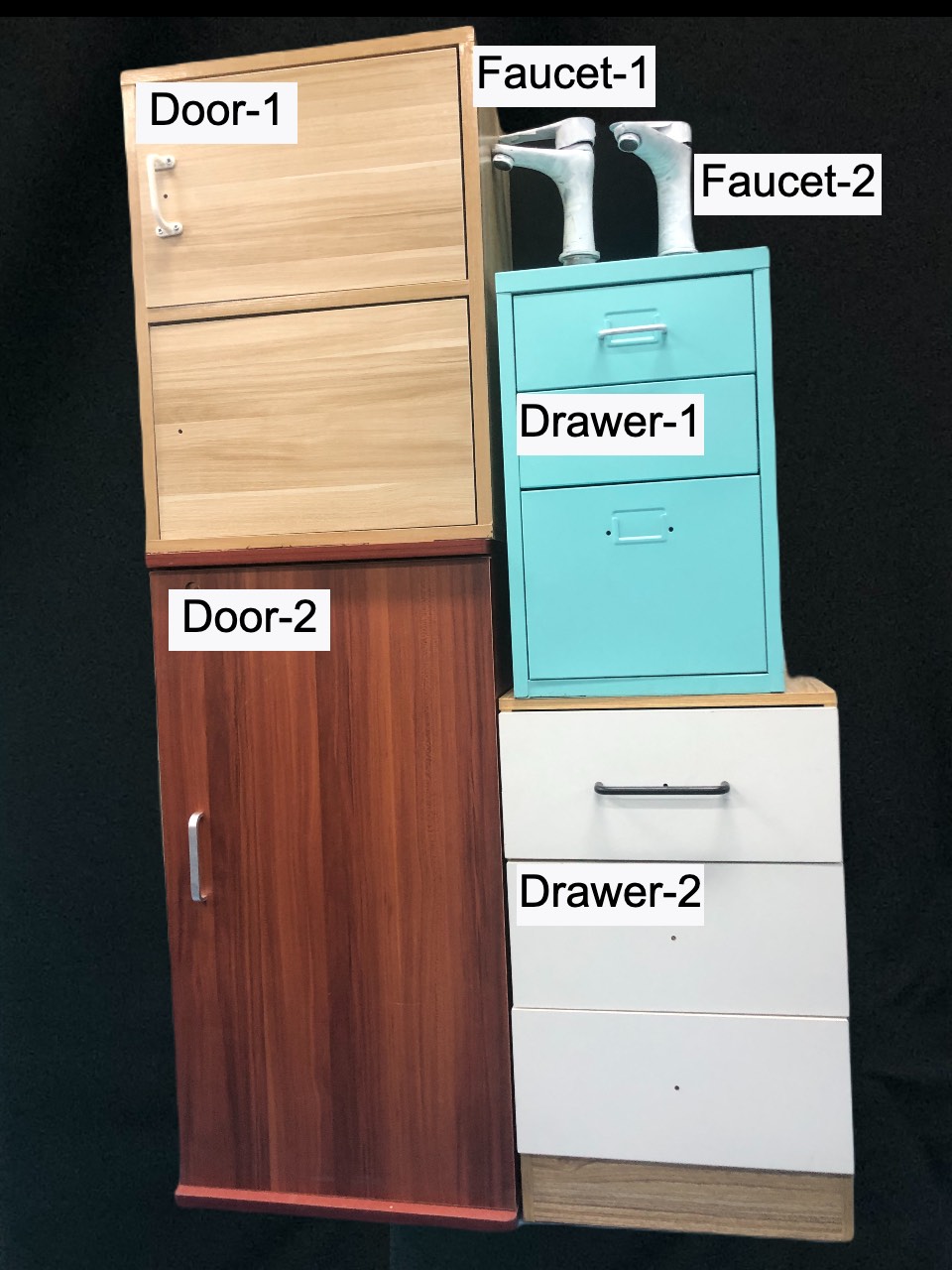}
    }
\caption{Real experimental setup and 3 categories of articulated objects for our experiment.}
\label{Fig.RealSetup}
\end{figure}

\subsection{Experimental Setup and \textcolor{black}{Evaluation Metrics}}
\label{Method.SetupImplementation}
\textbf{Experimental Setup}: As Fig. \ref{Fig.RealSetup} shows, our robot platform consists of an optical table, a 7-DOF robot (ROKAE xMate3Pro) with a 2-finger gripper (Robotiq 2F-140) mounted on its end link, an RGB-D camera (Intel RealSense D435). Besides, we randomly place the object in front of the robot arm within the robot's reachable workspace.

The simulation results presented in this study are obtained using the Sapien physics simulator \cite{xiang2020sapien}. We select three categories of articulated objects with a single movable part from the Partnet Mobility dataset \cite{xiang2020sapien} to collect synthetic data, including 40 doors, 16 drawers, and 14 faucets. In the RL policy training stage, 4 instances are selected from each category. \textcolor{black}{All methods are trained for 2 million steps, and we report the training results averaged across 7 independent experimental runs with randomized seeds.}

Our visual observation consists of a single RGB-D image captured from a hand-centric camera, with a size of $144 \times 256$. To facilitate the RL policy training with fast segmentation inference, we follow the streamlining design in Unet \cite{ronneberger2015u} and build it using MobileNetV2 \cite{sandler2018mobilenetv2} as encoders for its good feature representation and high efficiency. \textcolor{black}{Besides, the robot states include the joint angles, gripper finger position, and end-effector positions in the robot frame.} We set $K$, $T^{fc}$, and $K^{fc}$ to 4, 3, and 40. Moreover, \textcolor{black}{Soft Actor-Critic (SAC) \cite{haarnoja2018soft} is adopted as our RL algorithm.}

\textcolor{black}{\textbf{Evaluation Metrics}: We evaluate the performance of a policy by its success rate. Following related RL-based works \cite{urakami2019doorgym, qin2023dexpoint}, a task trial is considered a success if the movable part is moved at least a fixed range, typically set at 50\%. Manipulating the movable part to its full range poses a significant challenge due to the constrained workspace imposed by the fixed arm. In our work, we set the range 50\% for all tasks in simulation. In the real world, the range is 45$^{\circ}$ for doors and faucets, and 10 cm for drawers. In addition, we record the average success steps to evaluate the efficiency of a policy, which refers to the number of steps averaged by all successful trajectories.}

\begin{figure*}[tb]
\centering
\includegraphics[width=1\textwidth]{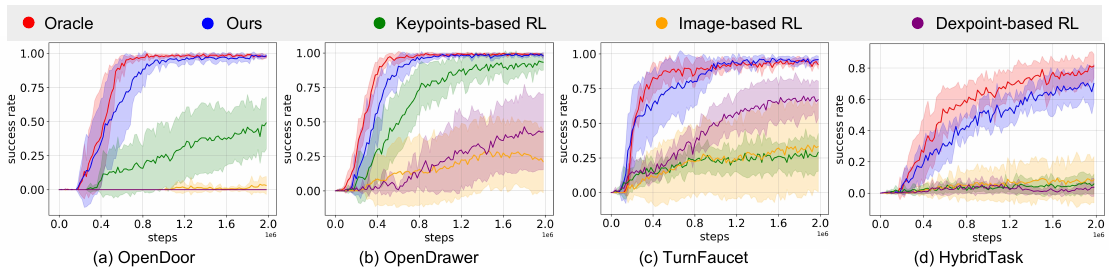}
\caption{\textcolor{black}{Comparsions of our method with baselines on three single tasks and the \textit{HybridTask} in simulation. The \textit{HybridTask} involves all three categories of articulated objects. The results are averaged over 7 random seeds.}}
\label{Fig.Baselines}
\end{figure*}

\begin{table*}[tbp]
\caption{The success rate of our method and baselines in simulation, averaged across 7 random seeds and 50 trials.}
\label{Tab.Evaluations}
\centering
\renewcommand\arraystretch{1.5}
\scalebox{1}{
\begin{tabular}{l|lccccc}
    \Xhline{1.5pt}
    \multicolumn{2}{l}{Methods}    & Image-based RL & Keypoints-based RL & Dexpoint-based RL & Oracle & \textbf{Ours} \\ 
    \Xhline{1pt}
    \multicolumn{2}{l}{OpenDoor}   & 0 $\pm$ 0     & 0.574$\pm$0.197 & 0 $\pm$ 0
    & 0.929 $\pm$ 0.062     & \textbf{0.871$\pm$0.06} \\
    \multicolumn{2}{l}{OpenDrawer} & 0.006$\pm$0.009 & 0.594$\pm$0.199 & 0.017 $\pm$ 0.042 & 0.914 $\pm$ 0.061     & \textbf{0.831$\pm$0.107} \\
    \multicolumn{2}{l}{TurnFaucet} & 0.294$\pm$0.182 & 0.246$\pm$0.168 & 0.243 $\pm$ 0.061 & 0.883 $\pm$ 0.045     & \textbf{0.843$\pm$0.083} \\
    \multirow{3}{*}{Hybrid} & Door & 0.029$\pm$0.018 & 0.040$\pm$0.049 & 0.006 $\pm$ 0.009 & 0,791 $\pm$ 0.079     & \textbf{0.726$\pm$0.102} \\
    \multirow{3}{*}{}   & Drawer   & 0.023$\pm$0.017 & 0.029$\pm$0.034 & 0.006 $\pm$ 0.009 & 0.857 $\pm$ 0.081     & \textbf{0.754$\pm$0.143} \\
    \multirow{3}{*}{}   & Faucet   & 0.080$\pm$0.043 & 0.009$\pm$0.015 & 0.009 $\pm$ 0.015 & 0.814 $\pm$ 0.081     & \textbf{0.737$\pm$0.132} \\
    \Xhline{1.5pt}
\end{tabular}}
\end{table*}

\subsection{Experiments in Simulation}
To evaluate the effectiveness of our proposed method for articulated object manipulation tasks, we compare it against several vision-based RL baselines. \textcolor{black}{Note that all baselines only differ in the visual representation, and all the other settings are kept the same.}

\begin{itemize}[leftmargin=0.3cm]
  \item \textcolor{black}{\textbf{Oracle}: This baseline directly obtains the ground-truth segmentation map from simulation, which is the upper bound of performance for our framework.}
  \item \textbf{Image-based RL}: \textcolor{black}{The segmentation map and the corresponding depth map are fed into an image encoder to extract visual features. We choose the image encoder from \cite{yarats2021image} due to its high efficiency for RL training. Our objective is to compare the part-guided image features with our part-guided geometric features.}
  \item \textbf{Keypoints-based RL}: \textcolor{black}{We follow \cite{wangLearningSemanticKeypoint2020a} and estimate 3 keypoints from handles. Then we put the keypoints with the robot states into the RL policy. Different from the original work, the keypoints are estimated from hand-centric RGB-D images during the manipulation process, introducing invisibility challenges. Despite this, we achieve comparable results (estimation error is around 1.6 cm without invisibility and about 2.1 cm for all scenarios).}
  \item \textcolor{black}{\textbf{Dexpoint-based RL}: We follow the same observation as \cite{qin2023dexpoint} in our hand-centric setting. Specifically, the filtered downsampled scene point clouds and the imagined points of our gripper are concatenated with their one-hot encoding as the PointNet input. The extracted geometric features and the robot states are then fed into the RL policy.}
\end{itemize}

We evaluate all methods on three single tasks and one \textit{HybridTask} that contains all three categories of articulated objects. For each task, we train the RL policy with one randomly selected object in each episode. To ensure a fair comparison, we evaluate the methods on one novel instance for each task and record their success rate over 50 trials.

\begin{figure}[bp] 
\centering
\includegraphics[width=0.48\textwidth]{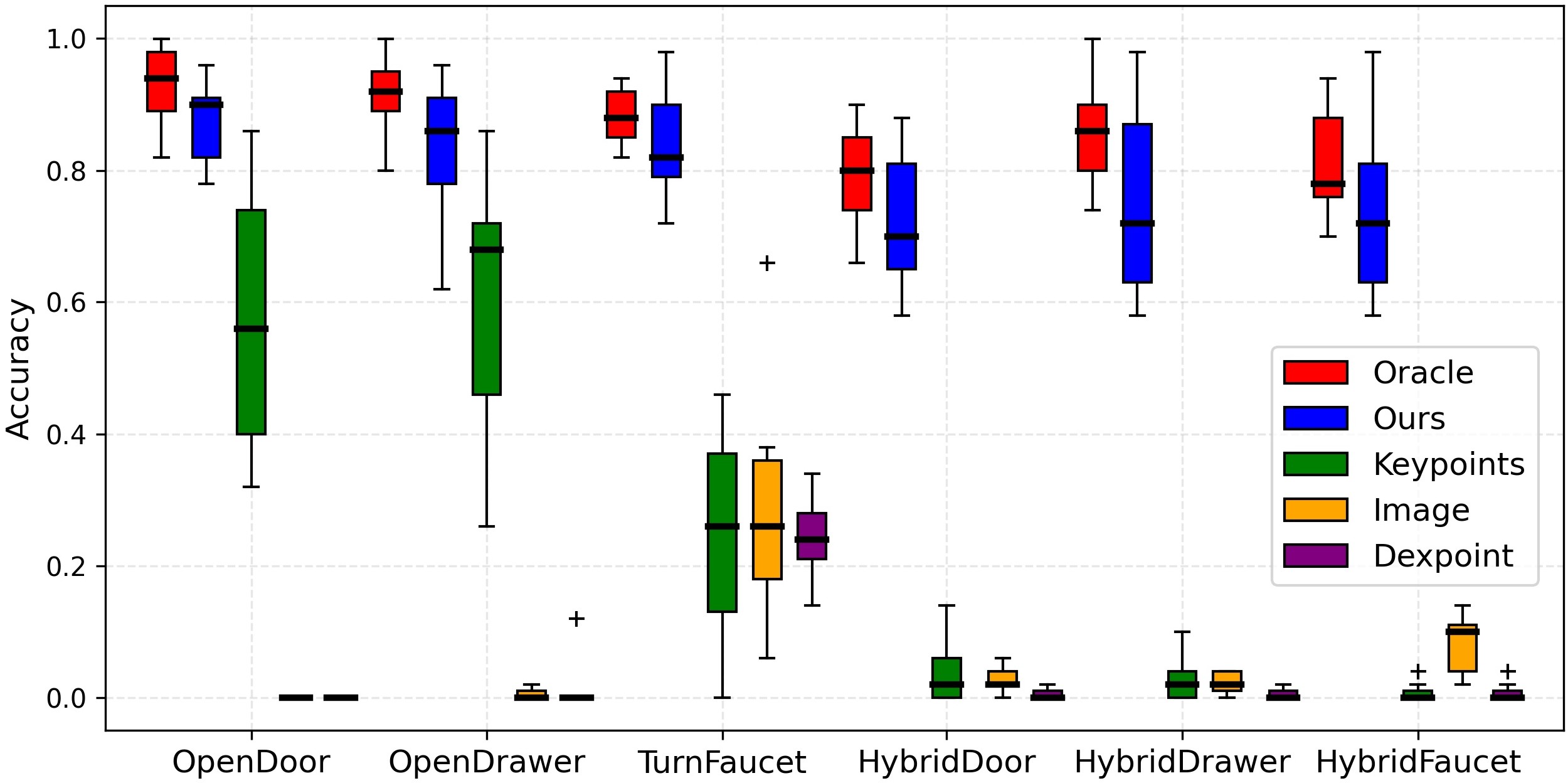}
\caption{\textcolor{black}{Success rates of our method and baselines in simulation, averaged across 7 random seeds. \textit{HybridDoor} represents the RL policy trained on the \textit{HybridTask} and evaluated on the \textit{OpenDoor} task. Similar meanings apply to \textit{HybridDrawer} and \textit{HybridFaucet}.}}
\label{Fig.EvalBaselines}
\end{figure}

As presented in Fig. \ref{Fig.Baselines}, our proposed framework demonstrates superior performance in terms of training efficiency and stability compared to the other baselines for all single manipulation tasks. \textcolor{black}{For the \textit{HybridTask}, our approach stands out as the sole method capable of concurrently learning from various types of articulated objects, attributing to our generalizable and condensed part representation.}

\textcolor{black}{Evaluation results is provided in Fig. \ref{Fig.EvalBaselines}, and the detailed results are shown in Table \ref{Tab.Evaluations}. The results indicate that our method generalizes well on novel instances, not only for single tasks but also for the \textit{HybridTask}. In contrast, the other baselines meet a significant performance drop facing novel instances.}

\begin{figure}[tbp] 
\centering
\includegraphics[width=0.35\textwidth]{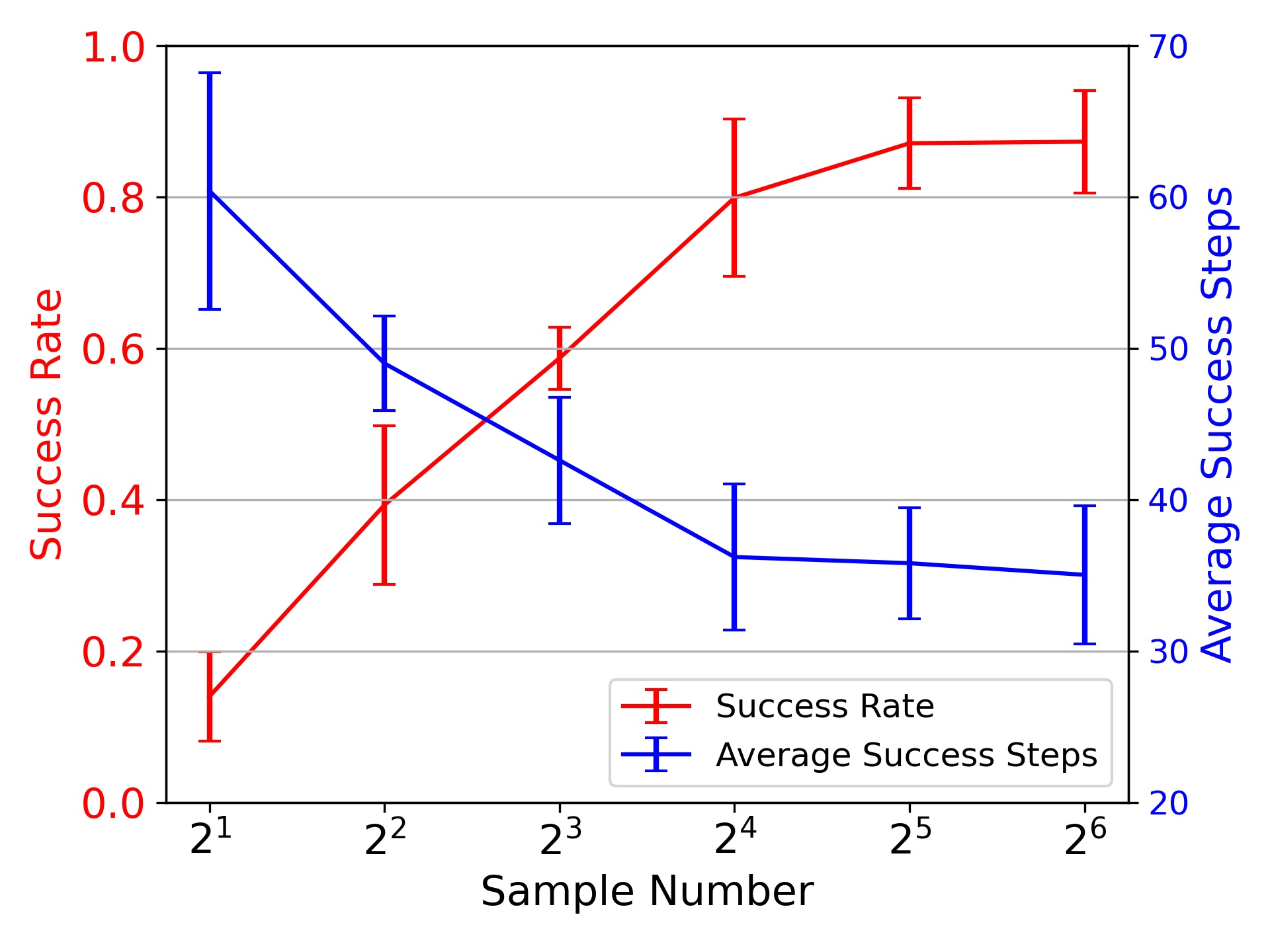}
\caption{\textcolor{black}{Success rate (Red) and average success steps (black) for different numbers of sampling points on the \textit{OpenDoor} task in simulation. The results are averaged over 7 random seeds.}}
\label{Fig.AblationSampleNum}
\end{figure}

\textbf{Ablation Study on Sampling Number}: To evaluate the impact of the sampling number $N_s$ on our manipulation policy, we conduct an ablation study on the \textit{OpenDoor} task. We train our method with different numbers of sampling points for each part while keeping all the other settings constant. The results are averaged over 7 random seeds.

In Fig. \ref{Fig.AblationSampleNum}, we present the results of our ablation experiment on the \textit{OpenDoor} task. \textcolor{black}{The performance of our method improves with an increasing number of sampling points and then reaches a plateau after a certain number of points. However, note that using more points also consumes additional computing resources. In our work, we set $N_s$ to 32 to strike a balance between performance and computational efficiency.}

\subsection{Sim-to-Real Transfer}
\textcolor{black}{We perform Sim2Real experiments to evaluate the performance of our proposed framework in the real world.}
\textbf{Part-guided Sampling Results}: Multiple sampling strategies based on the segmentation map are available. One common approach is Uniformly Downsampling used in 3D RL methods \cite{liu2022frame}. Accordingly, we downsample the scene point cloud to 1024 points in our experiments. Other strategies include Score-based Sampling, which selects the top $N_s$ points from each part based on their semantic scores; FPS, which samples the farthest $N_s$ points from each part; Random Sampling, which randomly samples $N_s$ points from each part; and our proposed FUS strategy.

\begin{figure}[t]
\centering
\includegraphics[width=0.4\textwidth]{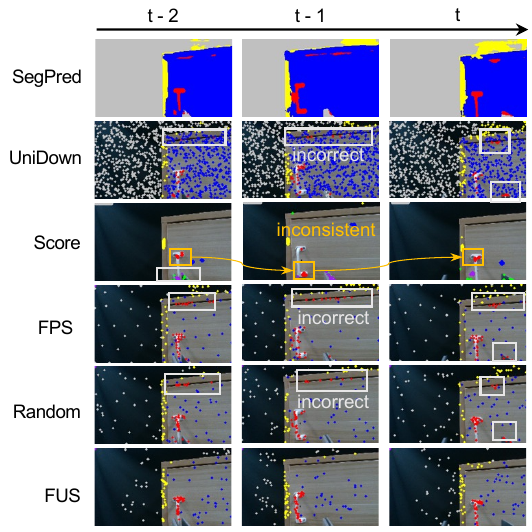}
\caption{The sampled points of different sampling methods on 3 consecutive real-world captured frames. The top row shows the segmentation predictions. The bottom rows show the sampled part points with different colors.}
\label{Fig.RealSampleablation}
\end{figure}

We compare various sampling methods and visualize the sampled points across three consecutive frames in a manipulation trajectory, as depicted in Fig. \ref{Fig.RealSampleablation}. The results indicate that uniformly-downsampled points may lack crucial points of small parts, such as handles, which can negatively affect the manipulation task. Score-based sampled points tend to be clustered in small circular regions, resulting in the loss of shape information of parts, and the positions of the sampled points vary rapidly across consecutive frames. Furthermore, none of the four alternative sampling methods, including FPS, Random Sampling, Score-based Sampling, and Uniformly Downsampling, can effectively handle segmentation errors and misclassified handle points, which can mislead the RL policy. In contrast, our proposed FUS strategy accurately and consistently samples points from each part, contributing to the stability of our RL policy.

\textbf{Manipulation Results}: To evaluate the effectiveness of our policy in the real world, we conduct trials on two doors, two drawers, and two faucets. These objects significantly vary in size, texture, and shape compared to the synthetic data. \textcolor{black}{We perform 20 trials for each object, recording the success rate and average success steps.} In each trial, we randomly position the object in front of the robot, ensuring the handle falls within the robot's operational space.

\begin{table}[tbp]
\caption{Success rate and average success steps in the real world}
\label{Tab.RealMani}
\centering
\renewcommand\arraystretch{1}
\begin{tabular}{l|ccc}
    \Xhline{1.5pt}
    \multicolumn{2}{c}{\textbf{Task}} & \textcolor{black}{\textbf{Dexpoint}} & \textbf{Ours} \\ 
    \Xhline{1pt}
    \multirow{3}{*}{\textbf{Single}} & OpenDoor & \makecell[c]{\textcolor{black}{0 / 40} \\ \textcolor{black}{N/A}} & \makecell[c]{\textbf{35 / 40} \\ $27.8\pm 3.2$} \\
    \multirow{3}{*}{}   & OpenDrawer & \makecell[c]{\textcolor{black}{0 / 40} \\ \textcolor{black}{N/A}}      & \makecell[c]{\textbf{32 / 40} \\ $28.3\pm 3.0$} \\
    \multirow{3}{*}{}   & TurnFaucet & \textcolor{black}{\makecell[c]{8 / 40 \\ $25.6\pm 3.6$}} & \makecell[c]{\textbf{35 / 40} \\ $19.9\pm 2.7$} \\
    \Xhline{1pt}
    \multirow{3}{*}{\textbf{Hybrid}} & OpenDoor & \makecell[c]{\textcolor{black}{0 / 40} \\ \textcolor{black}{N/A}}     & \makecell[c]{\textbf{31 / 40} \\ $31.1\pm 3.8$} \\
    \multirow{3}{*}{}   & OpenDrawer & \makecell[c]{\textcolor{black}{0 / 40} \\ \textcolor{black}{N/A}}     & \makecell[c]{\textbf{27 / 40} \\ $30.1\pm 4.2$} \\
    \multirow{3}{*}{}   & TurnFaucet & \makecell[c]{\textcolor{black}{0 / 40} \\ \textcolor{black}{N/A}}      & \makecell[c]{\textbf{32 / 40} \\ $19.7\pm 4.5$} \\
    \Xhline{1.5pt}
\end{tabular}
\end{table}

Table \ref{Tab.RealMani} reports the performance of our sampling-based 3D RL policy on the three articulated object manipulation tasks. Our framework achieves remarkable results on all three tasks. In particular, $RL_{hybrid}$, trained under the \textit{HybridTask}, demonstrates great performance on novel instances. \textcolor{black}{Despite utilizing the same domain randomization techniques, the Dexpoint-based RL method only attains a 20\% success rate on the \textit{TurnFaucet} task. This indicates that without specific \textcolor{black}{part segmentation} guidance, it faces challenges in manipulating articulated objects. Besides, both the Image-based RL and Keypoints-based RL methods exhibit a notable Sim2Real gap, resulting in failure to perform adequately in real-world scenarios.}

To evaluate the effectiveness of our sampling strategy, we conduct ablation experiments on the \textit{OpenDoor} task, and the results are presented in Table \ref{Tab.AblationSampleMode}. The results indicate that both uncertainty estimation and frame consistency contribute significantly to improving the performance of the framework.

\subsection{\textcolor{black}{Limitations and Failure Analysis}}
\textcolor{black}{\textbf{Limitations}: \textcolor{black}{Our work assumes predefined part specifications for different types of objects, restricting its generality.} Due to the hand-centric setting, we make the assumption that the handle can be seen in the initial frame. However, if the policy is misled by erroneous perception results and deviates from the correct path, it becomes challenging to relocate the handle. Additionally, our FUS strategy relies on the assumption that \textcolor{black}{the object's rigid base} remains stationary during the manipulation process, and it may fail if certain regions are consistently mis-segmented (e.g., the handle) in consecutive frames. Another limitation is that our method requires careful reward shaping to guide the learning process, which may not be easily extendable to more object types. Finally, as we introduce more types of articulated objects, our current training strategy becomes less effective and could benefit from improvements for more efficient training.}

\begin{figure}[tbp]
\centering
\includegraphics[width=0.4\textwidth]{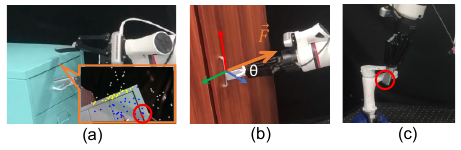}
\caption{\textbf{Failure Cases}: (a) Failing to approach due to incorrect sampled points; (b) Failing to pull due to inadequate force direction; (c) Failing to turn due to being stuck on the fixed link.}
\label{Fig.FailureCases}
\end{figure}

\begin{table}[tbp]
\caption{The success rate of different sampling strategies in the real world}
\label{Tab.AblationSampleMode}
\centering
\renewcommand\arraystretch{1.2}
{
\begin{tabular}{l c}
\Xhline{1.5pt}
Methods & Success Rate \\
\Xhline{1pt}
Random & 24 / 40 (60.0 \%)\\
\textcolor{black}{FPS} & \textcolor{black}{26 / 40 (65.0 \%)} \\
FUS \textit{w/o Uncertainty\quad\quad} & 29 / 40 (72.5 \%)\\
FUS \textit{w/o Consistency\quad\quad} & 31 / 40 (77.5 \%)\\
FUS & \textbf{35 / 40} (\textbf{87.5 \%})\\
\Xhline{1.5pt}
\end{tabular}}
\end{table}

\textbf{Failure Analysis}: Some of the failure cases are shown in Fig. \ref{Fig.FailureCases}. In case (a), despite generally reliable segmentation results, the policy can be misled by erroneous segmentation of background objects as handles. In case (b), we found that inadequate grasping is a major cause of failure in our framework. Improper grasp points often result in loose contact between the gripper and the handle, leading to suboptimal manipulation behavior. To tackle this issue, incorporating additional sensing modalities, such as force-torque sensors, can provide force feedback and enhance the policy's accuracy. In case (c), the robot gripper occasionally collided with the fixed part of the object, such as the faucet base. To address this, optimizing the placement of the gripper and the hand-centric camera could be beneficial.

\begin{figure}[tbp] 
\centering
\includegraphics[width=0.48\textwidth]{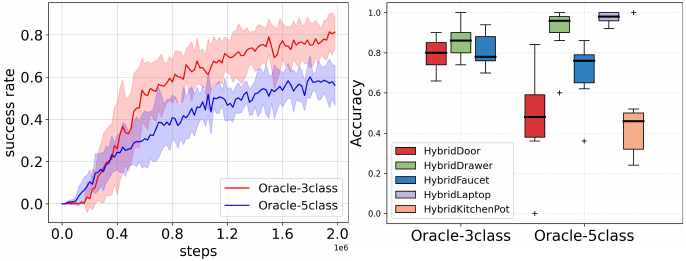}
\caption{\textcolor{black}{Success rate of our RL policy on different types of articulated objects in the simulation. The results are averaged over 7 random seeds.}}
\label{Fig.MultiArti}
\end{figure}

\textcolor{black}{\subsection{Analysis on the Versatile Policy}}
\textcolor{black}{To explore the versatility of our RL framework, we incorporate two more types of articulated objects: laptops and kitchen pots. \textcolor{black}{To exclude the influence of part segmentation errors, we employ oracle part segmentation in simulation to analyze the performance of the policy.} The training and evaluation results are depicted in Fig. \ref{Fig.MultiArti} and Table \ref{Tab.MultiArti}. On average, the success rate decreases by approximately 12\% (from 82.1\% to 70.3\%), considering all classes. Despite a certain decline in performance, the results prove the potential of our versatile RL policy for a broader range of articulated object manipulation tasks.}

\begin{table}[tbp]
\caption{The success rate of our method and baselines in simulation, averaged across 3 random seeds and 50 trials.}
\label{Tab.MultiArti}
\centering
\renewcommand\arraystretch{1.5}
\begin{tabular}{l|ccc}
    \Xhline{1.5pt}
    Methods    & Oracle-3class & Oracle-5class \\ 
    \Xhline{1pt}
    HybridDoor   & 0.791$\pm$0.079 & 0.466$\pm$0.243 &  \\
    HybridDrawer & 0.857$\pm$0.081 & 0.903$\pm$0.131 &  \\
    HybridFaucet & 0.814$\pm$0.081 & 0.694$\pm$0.155 &  \\
    HybridLaptop & N/A             & 0.974$\pm$0.028 &  \\
    HybridKitchenPot & N/A         & 0.477$\pm$0.234 &  \\
    \Xhline{1.5pt}
\end{tabular}
\end{table}

\section{CONCLUSION}
In this letter, we propose an efficient part-guided 3D RL framework for articulated object manipulation tasks. \textcolor{black}{The main contribution lies in the ability to learn a single versatile RL policy that can be applied across various tasks and directly deployed to novel real-world instances. Simulation and real-world experimental results demonstrate that our policy exhibits high accuracy and efficiency.}

\textcolor{black}{In future work, we plan to improve our part representation and extend our framework to a more general manipulation pipeline. In addition, we intend to refine the reward-tuning methods by adopting a more generalizable mechanism, which may minimize the design costs and make the overall process more efficient and adaptable.}

\bibliographystyle{IEEEtran}
\bibliography{IEEEabrv, Reference}

\begin{thebibliography}{10}
\providecommand{\url}[1]{#1}
\csname url@samestyle\endcsname
\providecommand{\newblock}{\relax}
\providecommand{\bibinfo}[2]{#2}
\providecommand{\BIBentrySTDinterwordspacing}{\spaceskip=0pt\relax}
\providecommand{\BIBentryALTinterwordstretchfactor}{4}
\providecommand{\BIBentryALTinterwordspacing}{\spaceskip=\fontdimen2\font plus
\BIBentryALTinterwordstretchfactor\fontdimen3\font minus \fontdimen4\font\relax}
\providecommand{\BIBforeignlanguage}[2]{{%
\expandafter\ifx\csname l@#1\endcsname\relax
\typeout{** WARNING: IEEEtran.bst: No hyphenation pattern has been}%
\typeout{** loaded for the language `#1'. Using the pattern for}%
\typeout{** the default language instead.}%
\else
\language=\csname l@#1\endcsname
\fi
#2}}
\providecommand{\BIBdecl}{\relax}
\BIBdecl

\bibitem{qin2020s4g}
Y.~Qin, R.~Chen, H.~Zhu, M.~Song \emph{et~al.}, ``S4g: Amodal single-view single-shot se(3) grasp detection in cluttered scenes,'' in \emph{Conference on robot learning}, 2020, pp. 53--65.

\bibitem{fang2020graspnet}
H.-S. Fang, C.~Wang, M.~Gou, and C.~Lu, ``Graspnet-1billion: A large-scale benchmark for general object grasping,'' in \emph{Proc. IEEE/CVF Conf. Comput. Vis. Pattern Recognit.}, 2020, pp. 11\,444--11\,453.

\bibitem{mo2021where2act}
K.~Mo, L.~J. Guibas, M.~Mukadam, A.~Gupta \emph{et~al.}, ``Where2act: From pixels to actions for articulated 3d objects,'' in \emph{Proc. IEEE Int. Conf. Comput. Vis.}, 2021, pp. 6813--6823.

\bibitem{wu2022vatmart}
R.~Wu, Y.~Zhao, K.~Mo, Z.~Guo \emph{et~al.}, ``{VAT}-mart: Learning visual action trajectory proposals for manipulating 3d {ART}iculated objects,'' in \emph{Proc. Int. Conf. Learn. Represent.}, 2022.

\bibitem{geng2022end}
Y.~Geng, B.~An, H.~Geng, Y.~Chen \emph{et~al.}, ``End-to-end affordance learning for robotic manipulation,'' \emph{arXiv preprint arXiv:2209.12941}, 2022.

\bibitem{xu2022universal}
Z.~Xu, Z.~He, and S.~Song, ``Universal manipulation policy network for articulated objects,'' \emph{{IEEE} Robot. Autom. Lett.}, vol.~7, no.~2, pp. 2447--2454, 2022.

\bibitem{EisnerZhang2022FLOW}
B.~Eisner*, H.~Zhang*, and D.~Held, ``Flowbot3d: Learning 3d articulation flow to manipulate articulated objects,'' in \emph{Robotics: Science and Systems (RSS)}, 2022.

\bibitem{ma2023sim2real2}
L.~Ma, J.~Meng, S.~Liu, W.~Chen \emph{et~al.}, ``Sim2real2: Actively building explicit physics model for precise articulated object manipulation,'' in \emph{Proc. IEEE Int. Conf. Robot. Autom.}, 2023, pp. 11\,698--11\,704.

\bibitem{jiang2022ditto}
Z.~Jiang, C.-C. Hsu, and Y.~Zhu, ``Ditto: Building digital twins of articulated objects from interaction,'' in \emph{Proc. IEEE/CVF Conf. Comput. Vis. Pattern Recognit.}, 2022, pp. 5616--5626.

\bibitem{wang2022adaafford}
Y.~Wang, R.~Wu, K.~Mo, J.~Ke \emph{et~al.}, ``Adaafford: Learning to adapt manipulation affordance for 3d articulated objects via few-shot interactions,'' in \emph{Proc. Eur. Conf. Comput. Vis.}, 2022, pp. 90--107.

\bibitem{urakami2019doorgym}
Y.~Urakami, A.~Hodgkinson, C.~Carlin, R.~Leu \emph{et~al.}, ``Doorgym: A scalable door opening environment and baseline agent,'' \emph{arXiv preprint arXiv:1908.01887}, 2019.

\bibitem{wangLearningSemanticKeypoint2020a}
J.~Wang, S.~Lin, C.~Hu, Y.~Zhu \emph{et~al.}, ``Learning semantic keypoint representations for door opening manipulation,'' \emph{{IEEE} Robot. Autom. Lett.}, vol.~5, no.~4, pp. 6980--6987, 2020.

\bibitem{yen2020learning}
L.~Yen-Chen, A.~Zeng, S.~Song, P.~Isola \emph{et~al.}, ``Learning to see before learning to act: Visual pre-training for manipulation,'' in \emph{Proc. IEEE Int. Conf. Robot. Autom.}, 2020, pp. 7286--7293.

\bibitem{radosavovic2022realworld}
I.~Radosavovic, T.~Xiao, S.~James, P.~Abbeel \emph{et~al.}, ``Real-world robot learning with masked visual pre-training,'' in \emph{Conference on Robot Learning}, 2023, pp. 416--426.

\bibitem{seo2022reinforcement}
Y.~Seo, K.~Lee, S.~L. James, and P.~Abbeel, ``Reinforcement learning with action-free pre-training from videos,'' in \emph{Proc. Int. Conf. Mach. Learn.}, 2022, pp. 19\,561--19\,579.

\bibitem{zhan2022learning}
A.~Zhan, R.~Zhao, L.~Pinto, P.~Abbeel \emph{et~al.}, ``Learning visual robotic control efficiently with contrastive pre-training and data augmentation,'' in \emph{Proc. IEEE/RSJ Int. Conf. Intell. Rob. Syst.}, 2022, pp. 4040--4047.

\bibitem{ma2023vip}
Y.~J. Ma, S.~Sodhani, D.~Jayaraman, O.~Bastani \emph{et~al.}, ``{VIP}: Towards universal visual reward and representation via value-implicit pre-training,'' in \emph{Proc. Int. Conf. Learn. Represent.}, 2023.

\bibitem{xiao2022maskedvpt}
T.~Xiao, I.~Radosavovic, T.~Darrell, and J.~Malik, ``Masked visual pre-training for motor control,'' \emph{arXiv preprint arXiv:2203.06173}, 2022.

\bibitem{kaelbling1998planning}
L.~P. Kaelbling, M.~L. Littman, and A.~R. Cassandra, ``Planning and acting in partially observable stochastic domains,'' \emph{Artificial intelligence}, vol. 101, no. 1-2, pp. 99--134, 1998.

\bibitem{hsu2022visionbased}
K.~Hsu, M.~J. Kim, R.~Rafailov, J.~Wu \emph{et~al.}, ``Vision-based manipulators need to also see from their hands,'' \emph{arXiv preprint arXiv:2203.12677}, 2022.

\bibitem{geng2023gapartnet}
H.~Geng, H.~Xu, C.~Zhao, C.~Xu \emph{et~al.}, ``Gapartnet: Cross-category domain-generalizable object perception and manipulation via generalizable and actionable parts,'' in \emph{Proc. IEEE/CVF Conf. Comput. Vis. Pattern Recognit.}, 2023, pp. 7081--7091.

\bibitem{tobin2017domain}
J.~Tobin, R.~Fong, A.~Ray, J.~Schneider \emph{et~al.}, ``Domain randomization for transferring deep neural networks from simulation to the real world,'' in \emph{Proc. IEEE/RSJ Int. Conf. Intell. Rob. Syst.}, 2017, pp. 23--30.

\bibitem{song2015sun}
S.~Song, S.~P. Lichtenberg, and J.~Xiao, ``Sun rgb-d: A rgb-d scene understanding benchmark suite,'' in \emph{Proc. IEEE/CVF Conf. Comput. Vis. Pattern Recognit.}, 2015, pp. 567--576.

\bibitem{mccormac2017scenenet}
J.~McCormac, A.~Handa, S.~Leutenegger, and A.~J. Davison, ``Scenenet rgb-d: Can 5m synthetic images beat generic imagenet pre-training on indoor segmentation?'' in \emph{Proc. IEEE Int. Conf. Comput. Vis.}, 2017, pp. 2678--2687.

\bibitem{zhang2022close}
X.~Zhang, R.~Chen, A.~Li, F.~Xiang \emph{et~al.}, ``Close the optical sensing domain gap by physics-grounded active stereo sensor simulation,'' \emph{{IEEE} Trans. Robot.}, vol.~39, no.~3, pp. 2429--2447, 2023.

\bibitem{yarats2021image}
D.~Yarats, I.~Kostrikov, and R.~Fergus, ``Image augmentation is all you need: Regularizing deep reinforcement learning from pixels,'' in \emph{International conference on learning representations}, 2020.

\bibitem{liu2022frame}
M.~Liu, X.~Li, Z.~Ling, Y.~Li \emph{et~al.}, ``Frame mining: a free lunch for learning robotic manipulation from 3d point clouds,'' in \emph{Conference on Robot Learning}, 2022.

\bibitem{qin2023dexpoint}
Y.~Qin, B.~Huang, Z.-H. Yin, H.~Su \emph{et~al.}, ``Dexpoint: Generalizable point cloud reinforcement learning for sim-to-real dexterous manipulation,'' in \emph{Conference on Robot Learning}, 2023, pp. 594--605.

\bibitem{qi2019deep}
C.~R. Qi, O.~Litany, K.~He, and L.~J. Guibas, ``Deep hough voting for 3d object detection in point clouds,'' in \emph{Proc. IEEE Int. Conf. Comput. Vis.}, 2019, pp. 9277--9286.

\bibitem{kendall2017uncertainties}
A.~Kendall and Y.~Gal, ``What uncertainties do we need in bayesian deep learning for computer vision?'' in \emph{Proc. Adv. Neural Inf. Process. Syst.}, vol.~30, 2017.

\bibitem{wang2019aleatoric}
G.~Wang, W.~Li, M.~Aertsen, J.~Deprest \emph{et~al.}, ``Aleatoric uncertainty estimation with test-time augmentation for medical image segmentation with convolutional neural networks,'' \emph{Neurocomputing}, vol. 338, pp. 34--45, 2019.

\bibitem{qi2017pointnet}
C.~R. Qi, H.~Su, K.~Mo, and L.~J. Guibas, ``Pointnet: Deep learning on point sets for 3d classification and segmentation,'' in \emph{Proc. IEEE/CVF Conf. Comput. Vis. Pattern Recognit.}, 2017, pp. 652--660.

\bibitem{xiang2020sapien}
F.~Xiang, Y.~Qin, K.~Mo, Y.~Xia \emph{et~al.}, ``Sapien: A simulated part-based interactive environment,'' in \emph{Proc. IEEE/CVF Conf. Comput. Vis. Pattern Recognit.}, 2020, pp. 11\,097--11\,107.

\bibitem{ronneberger2015u}
O.~Ronneberger, P.~Fischer, and T.~Brox, ``U-net: Convolutional networks for biomedical image segmentation,'' in \emph{Proc. Int. Conf. Med. Image Comput. Comput.-Assisted Intervention}, 2015, pp. Part III 18 234--241.

\bibitem{sandler2018mobilenetv2}
M.~Sandler, A.~Howard, M.~Zhu, A.~Zhmoginov \emph{et~al.}, ``Mobilenetv2: Inverted residuals and linear bottlenecks,'' in \emph{Proc. IEEE/CVF Conf. Comput. Vis. Pattern Recognit.}, 2018, pp. 4510--4520.

\bibitem{haarnoja2018soft}
T.~Haarnoja, A.~Zhou, P.~Abbeel, and S.~Levine, ``Soft actor-critic: Off-policy maximum entropy deep reinforcement learning with a stochastic actor,'' in \emph{Proc. Int. Conf. Mach. Learn.}, 2018, pp. 1861--1870.

\end{thebibliography}

\addtolength{\textheight}{-12cm}   

\end{document}